\documentclass[letterpaper, 10 pt, conference]{ieeeconf}  

\IEEEoverridecommandlockouts                              

\let\labelindent\relax

\usepackage[numbers]{natbib}
\usepackage{multicol}
\usepackage[bookmarks=true]{hyperref}
\usepackage{amsmath}
\usepackage{amssymb}
\usepackage{color}
\usepackage{graphicx}
\usepackage{appendix}
\graphicspath{{images/}}

\usepackage{xspace}

\usepackage{setspace}

\usepackage{amsmath}
\usepackage{amsfonts}
\usepackage{amssymb}
\usepackage{amsthm}
\usepackage{bm}
\usepackage{bbm}
\usepackage{mathtools}
\usepackage{enumitem}
\usepackage{thmtools,thm-restate}
\usepackage{algorithm}
\usepackage{algorithmic}
\usepackage{color}
\usepackage{graphicx}
\usepackage{comment}

\def\Rbb{\mathbb{R}}

\def\R{\Rbb}

\def\*{\star}

\newcommand{\tr}[1]{ \mathrm{tr}\left( #1\right)}

\usepackage[dvipsnames]{xcolor}

\renewcommand{\tr}{{T}}

\newcommand{\Lag}{\mathcal{L}}
\newcommand{\Ham}{\mathcal{H}}

\newcommand{\p}{\mathbf{p}}
\newcommand{\q}{\mathbf{q}}
\newcommand{\qd}{{\dot{\q}}}

\newcommand{\uu}{\mathbf{u}}

\newcommand{\vv}{\mathbf{v}}

\newcommand{\x}{\mathbf{x}}
\newcommand{\xd}{{\dot{\x}}}
\newcommand{\xdd}{{\ddot{\x}}}

\newcommand{\z}{\mathbf{z}}

\newcommand{\f}{\mathbf{f}}
\newcommand{\h}{\mathbf{h}}

\newcommand{\zero}{\mathbf{0}}

\newcommand{\G}{\mathbf{G}}

\newcommand{\I}{\mathbf{I}}

\newcommand{\M}{\mathbf{M}}

\newcommand{\mP}{\mathbf{P}}
\newcommand{\mR}{\mathbf{R}}

\usepackage{amsmath}  
\usepackage{amssymb}
\usepackage{accents}  
\usepackage{amsthm}

\theoremstyle{plain}
\newtheorem{theorem}{Theorem}[section]
\newtheorem{lemma}[theorem]{Lemma}

\theoremstyle{definition}

\theoremstyle{remark}

\makeatletter
\let\save@mathaccent\mathaccent
\newcommand*\if@single[3]{%
  \setbox0\hbox{${\mathaccent"0362{#1}}^H$}%
  \setbox2\hbox{${\mathaccent"0362{\kern0pt#1}}^H$}%
  \ifdim\ht0=\ht2 #3\else #2\fi
  }
\newcommand*\rel@kern[1]{\kern#1\dimexpr\macc@kerna}
\newcommand*\widebar[1]{\@ifnextchar^{{\wide@bar{#1}{0}}}{\wide@bar{#1}{1}}}
\newcommand*\wide@bar[2]{\if@single{#1}{\wide@bar@{#1}{#2}{1}}{\wide@bar@{#1}{#2}{2}}}
\newcommand*\wide@bar@[3]{%
  \begingroup
  \def\mathaccent##1##2{%
    \let\mathaccent\save@mathaccent
    \if#32 \let\macc@nucleus\first@char \fi
    \setbox\z@\hbox{$\macc@style{\macc@nucleus}_{}$}%
    \setbox\tw@\hbox{$\macc@style{\macc@nucleus}{}_{}$}%
    \dimen@\wd\tw@
    \advance\dimen@-\wd\z@
    \divide\dimen@ 3
    \@tempdima\wd\tw@
    \advance\@tempdima-\scriptspace
    \divide\@tempdima 10
    \advance\dimen@-\@tempdima
    \ifdim\dimen@>\z@ \dimen@0pt\fi
    \rel@kern{0.6}\kern-\dimen@
    \if#31
      \overline{\rel@kern{-0.6}\kern\dimen@\macc@nucleus\rel@kern{0.4}\kern\dimen@}%
      \advance\dimen@0.4\dimexpr\macc@kerna
      \let\final@kern#2%
      \ifdim\dimen@<\z@ \let\final@kern1\fi
      \if\final@kern1 \kern-\dimen@\fi
    \else
      \overline{\rel@kern{-0.6}\kern\dimen@#1}%
    \fi
  }%
  \macc@depth\@ne
  \let\math@bgroup\@empty \let\math@egroup\macc@set@skewchar
  \mathsurround\z@ \frozen@everymath{\mathgroup\macc@group\relax}%
  \macc@set@skewchar\relax
  \let\mathaccentV\macc@nested@a
  \if#31
    \macc@nested@a\relax111{#1}%
  \else
    \def\gobble@till@marker##1\endmarker{}%
    \futurelet\first@char\gobble@till@marker#1\endmarker
    \ifcat\noexpand\first@char A\else
      \def\first@char{}%
    \fi
    \macc@nested@a\relax111{\first@char}%
  \fi
  \endgroup
}
\makeatother

\newcommand{\fullversiononly}{\iftrue}
\newcommand{\compressedversiononly}{\iffalse}

\title{\LARGE \bf
Generalized Nonlinear and Finsler Geometry for Robotics
}
\author{Nathan D. Ratliff$^{1}$, Karl Van Wyk$^{1}$, Mandy Xie$^{1,2}$, Anqi Li$^{1,3}$, and Muhammad Asif Rana$^{1,2}$
\thanks{$^{1}$All authors are with (or interned at) NVIDIA, 
$^{2}$Georgia Tech, $^{3}$University of Washington}%
}

\begin{document}

\maketitle
\thispagestyle{empty}
\pagestyle{empty}

\begin{abstract}
Robotics research has found numerous important applications of Riemannian geometry. Despite that, the concept remain challenging to many roboticists because the background material is  complex and strikingly foreign. Beyond {\em Riemannian} geometry, there are many natural generalizations in the mathematical literature---areas such as Finsler geometry and spray geometry---but those generalizations are largely inaccessible, and as a result there remain few applications within robotics. This paper presents a re-derivation of spray and Finsler geometries we found critical for the development of our recent work on a powerful behavioral design tool we call geometric fabrics. These derivations build from basic tools in advanced calculus and the calculus of variations making them more accessible to a robotics audience than standard presentations. We focus on the pragmatic and calculable results, avoiding the use of tensor notation to appeal to a broader audience, emphasizing geometric {\em path consistency} over ideas around connections and curvature. We hope that these derivations will contribute to an increased understanding of generalized nonlinear, and even classical Riemannian, geometry within the robotics community and inspire future research into new applications.
\end{abstract}

\section{Introduction}

Nonlinear geometry is in many ways fundamental to robotics. Robotic configuration spaces are naturally modeled as manifolds \cite{MurrayLiSastry94}, the classical mechanical dynamics of the robot is intimately linked to a Riemannian geometry \cite{bullo2004geometric}, Gauss-Newton optimization has strong application in vision \cite{TannerDART2015} and motion optimization \cite{KOMOToussaint2014} and those algorithms are closely related to Riemannian geometry \cite{RIEMORatliff2015ICRA,AbsilOptimizationOnManifolds2008} natural gradients \cite{AmariInformationGeometry1994} which are core to many modern machine learning methods \cite{pmlr-v37-schulman15}. Despite their importance, however, even classical Riemannian geometry remains inaccessible to many roboticists. And beyond that, Finsler \cite{baoRiemanFinslerGeometry2000} and spray geometry \cite{shenSprayGeometry2000}, natural generalizations of Riemannian, are largely unheard of within the community. How many applications are out-of-reach as a result?

Our own work on Riemannian Motion Policies (RMPs) \cite{ratliff2018rmps,cheng2018rmpflow} has led us to a study of what we call geometric fabrics \cite{geometricFabricsForMotionArXiv2020,optimizationFabricsForBehavioralDesignArXiv2020} as a formal provably stable model of reactive behavior, and to get there, we had to address this gap. Both Finsler and spray geometries proved critical for the development of that work, but were largely confined within opaque mathematical manuscripts mired with abstract concepts and foreign notation. The importance of the material, however, led us to entirely re-derive the foundations, rigorously, but in a language we could understand. This paper presents these re-derivations in a hope that they will be more accessible to a broader robotics audience and inspire new future applications.

We use notations of advanced calculus \cite{follandAdvancedCalculus2001} as much as possible and build on the Calculus of Variations \cite{CalculusOfVariationsGelfandFomin63} (Finsler geometry maybe viewed as the geometry of these functional optima), and we draw connections to classical mechanics \cite{ClassicalMechanicsTaylor05}, Riemannian geometry \cite{LeeRiemannianManifolds97,bullo2004geometric}, and the interrelations observed in geometric mechanics \cite{bullo2004geometric}. We limit ourselves to a study of {\em geometric path consistency} which we call generalized nonlinear geometry (see Section~\ref{sec:GeneralNonlinearGeometry}),\footnote{Our terminology differs from the literature. Equations exhibiting geometric path consistency are commonly called sprays in mathematics \cite{shenSprayGeometry2000}, and literature further develops more general geometry with a very similar name (a semi-spray), which doesn't exhibit such path consistency. We, therefore, restrict the term {\em geometry} to imply this concrete notion of {\em path consistency} to aid intuition and reserve the term semi-spray in our applications \cite{geometricFabricsForMotionArXiv2020} for these more general non-path-consistent equations.} and show that Finsler geometries are a type of generalized nonlinear geometry (see Section~\ref{sec:FinslerGeometry}), inheriting their geometric consistency.

\begin{figure}[!t]
   \vspace{2.5mm}
   \centering
   \compressedversiononly
   \includegraphics[width=.7\linewidth]{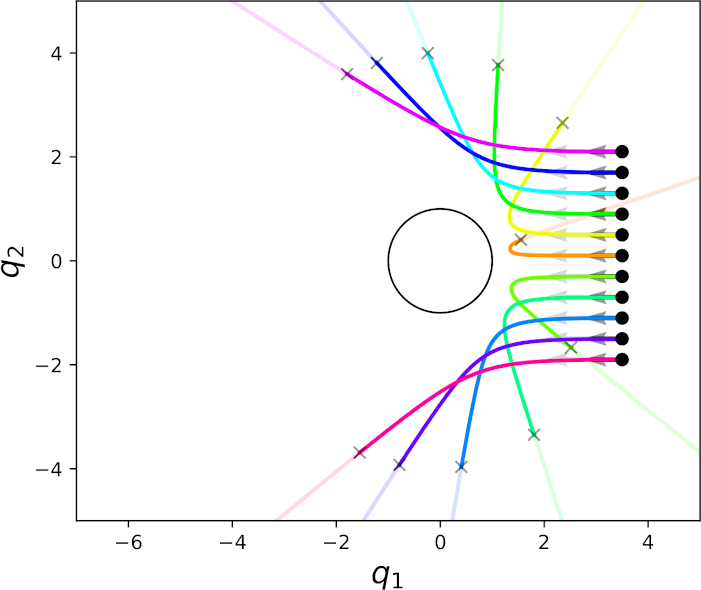}
   \else
   \includegraphics[width=.95\linewidth]{nonlinear_geometry/figures/geometric_consistency.jpg}
   \fi
   \caption{Paths generated from particle motion initiated at different speeds. Faded paths indicate higher $\alpha$ and opaque paths indicate lower $\alpha$.}
   \label{fig:geometric_consistency}
   \vspace{-2.5mm}
\end{figure}

Manifolds \cite{LeeSmoothManifolds2012} are a standard mathematical foundation of modern nonlinear geometry. However, these concepts can be very abstract and daunting for those unfamiliar. We, therefore, present our derivations in specific coordinates, following the conventions of many texts on classical mechanics \cite{ClassicalMechanicsTaylor05}. For those familiar with the concepts, we can say we do so without loss of generality---the Euler-Lagrange equation is naturally covariant, enabling curvilinear changes of coordinates as necessary while traversing a manifold with no change to the underlying behavior; covariant behavior of more general nonlinear geometries can be expressed through a transform tree as was done in \cite{cheng2018rmpflow} in the definition of {\em structured} Geometric Dynamical Systems (GDS).

\compressedversiononly
Some theoretical details have been omitted in this manuscript due to space restrictions, but can be found in the extended version of the paper \cite{finslerGeometryForRoboticsArXiv2020}.\footnote{See also {\scriptsize \url{ https://sites.google.com/nvidia.com/geometric-fabrics}} for videos and other relevant work.}
\fi 

\section{Generalized nonlinear geometries} \label{sec:GeneralNonlinearGeometry}

A generalized nonlinear geometry (known as a spray in mathematics \cite{shenSprayGeometry2000}) is a second-order differential equation describing a smooth collection of paths. These paths are equivalence classes of trajectories all passing through the same points but differing in speed profile. In essence, we allow a trajectory to speed up or slow down arbitrarily, and as long as its geometric shape remains consistent (i.e. it passes through the same one-dimensional set (submanifold) of points in space) we say the trajectory follows the same path (i.e. is part of the equivalence class). Colloquially, similar to how we can think of an arbitrary second-order differential equation as a collection of trajectories (its integral curves), we can think of the geometry as a collection of tubes. Each tube represents a path and contains multiple trajectories (infinitely many of them), the collection of all trajectories following that path with differing speed profiles (see Figure~\ref{fig:geometric_consistency}).

Concretely, two trajectories are said to be equivalent, and hence along the same path, if they are a {\em time reparameterization} away from one another. Given a trajectory with time index $t$ denoted $\x_t(t)$, a time reparameterization is a smooth, strictly monotonically increasing, nonlinear function $t:\R\rightarrow\R$ denoted $t(s)$ giving rise to a new time index $s$. The time reparameterization creates a new trajectory $\x_s(s) = \x_t\big(t(s)\big)$. Since, for a given $s_0$ and corresponding $t_0 = t(s_0)$, the points of the trajectories align
\begin{align}
    \x_s(s_0) = \x_t\big(t(s_0)\big) = \x_t(t_0),
\end{align}
and since this relationship is a bijection (due to the strict monotonicity),\footnote{Formally, it is a diffeomorphism between coordinate charts of the same one-dimensional manifold of points.} the two trajectories $\x_s(s)$ and $\x_t(t)$ pass through the same set of points and we say they follow the same path. Since $\frac{d}{ds}\x_t\big(t(s)\big) = \frac{d\x_t}{dt}\frac{dt}{ds}$
and $\frac{d^2}{ds^2}\x_t\big(t(s)\big) = \frac{d^2\x_t}{dt^2}\left(\frac{dt}{ds}\right)^2 + \frac{d\x_t}{dt}\frac{d^2t}{ds^2}$, we see velocities and accelerations under the time reparameterization are linked to one another as
\begin{align}\label{eqn:TimeReparamDerivs}
    \xd_s &= \frac{dt}{ds}\xd_t \\\nonumber
    \xdd_s &= \left(\frac{dt}{ds}\right)^2\xdd_t + \frac{d^2t}{ds^2}\xd_t,
\end{align}
where notationally the dots are understood to be time derivatives w.r.t. their respective time indices. E.g. $\xd_s = \frac{d\x_s}{ds}$, $\xd_t = \frac{d\x_t}{dt}$ and so forth. A smooth nonlinear geometry is defined by the collection of all time-reparameterization invariant smooth paths in a space. For every point $\x_0$ and speed-independent direction vector $\widehat{\vv}$, the nonlinear geometry defines a unique path eminating from that point $\x_0$ following the specified initial direction $\widehat{\vv}$.

Denoting the orthogonal projector projecting orthogonally to velocity as $\mP_\xd^\perp = \I - \widehat{\xd}\widehat{\xd}^\tr$, the family of geometries we consider here are those characterized by a second-order differential equation of the form
\begin{align}
    \mP_\xd^\perp\Big[\xdd + \h_2(\x, \xd)\Big] = \zero,
\end{align}
where $\h_2(\x, \xd)$ is a smooth function that is {\em positively homogeneous of degree 2} (HD2) in velocities.\footnote{Generally, a function $f(\z)$ is said to be positively homogeneous of degree $k$ (abbreviated HD$k$) if $f(\lambda\z) = \lambda^kf(\z)$ for $\lambda\geq0$ \cite{lewisHomogeneousFunctions1969}. Homogeneity of degree 1 and 2 is used in the definitions below as well. In this case, $\h_2$ must be HD2, meaning $\h_2(\x, \lambda\xd) = \lambda^2\h_2(\x, \xd)$ for $\lambda>0$.} We call equations of this form {\em geometric equations}.

Since the projector $\mP_\xd^\perp$ is reduced rank, there is solution redundancy. We will see that this redundancy precisely describes the ability to arbitrarily speed up or slow down along a trajectory while sticking to the same path.

The interior equation in isolation
\begin{align}
    \xdd + \h_2(\x, \xd) = \zero
\end{align}
we call the {\em generating equation} and is said to {\em generate} the geometry via its system of trajectories. Note that since $\mP_\xd^\perp$ has null space spanned by $\xd$, solutions to the geometric equation are solutions to 
\begin{align}
    \xdd + \h_2(\x, \xd) + \alpha(t)\xd = \zero
\end{align}
where $\alpha:\R\rightarrow\R$ is a smooth function defining an acceleration along the direction of motion. We call this equation the {\em explicit form geometric equation}.

\begin{theorem}
All time reparameterizations of generating solutions are geometric solutions, and each geometric solution characterized by starting position $\x_0$ and direction $\widehat{\vv}$ is a time reparameterization away from any generating solution with initial conditions $(\x_0, \gamma\widehat{\vv})$ for some $\gamma > 0$.
\end{theorem}
\fullversiononly
\begin{proof}
We first address reparameterization of generating solutions. Let $\x_s(s)$ be a generating solution trajectory and let $s(t)$ be an arbitrary time reparameterization. In terms of its inverse $t(s)$ (which always exists since $s(t)$ is strictly monotonically increasing by definition), by Equations~\ref{eqn:TimeReparamDerivs} we have
\begin{align}
    &\ \ \ \xdd_s + \h_2(\x_s, \xd_s) = \zero \\
    &\Rightarrow \left(\frac{dt}{ds}\right)^2\xdd_t + \frac{d^2t}{ds^2}\xd_t + \h_2(\x_t, \frac{dt}{ds}\xd_t) = \zero \\
    &\Rightarrow \left(\frac{dt}{ds}\right)^2\big[\xdd_t + \h_2(\x_t, \xd_t)\big]
        + \frac{d^2t}{ds^2}\xd_t = \zero \\
    &\Rightarrow \xdd_t + \h_2(\x_t, \xd_t) + \alpha\xd_t = \zero,
\end{align}
where $\alpha = \left(\frac{dt}{ds}\right)^{-2}\frac{d^2t}{ds^2}$. Thus, $\x_t(t)$ solves an explicit form geometric equation and is, therefore, a geometric solution.

Next, let $\x_s(s)$ be any solution to the geometric equation. Then at every point,
\begin{align}
    \xdd_s + \h_2(\x_s, \xd_s) + \alpha(s)\xd_s = \zero
\end{align}
for some smooth function $\alpha(s)$ across the trajectory. Under a time reparameterization $s(t)$ we get the following equation in terms of $\x_t$
\begin{align}
    &\left(\frac{dt}{ds}\right)^2\xdd_t + \frac{d^2t}{ds^2}\xd_t
    + \h_2(\x_t, \frac{dt}{ds}\xd_t)\\ \nonumber
    &\ \ \ \ \ \ \ \ \ \ \ \ \ \ \ \ \ \ \ \ \ \ \ \ \
    + \alpha(s) \frac{dt}{ds}\xd_t = \zero \\
    &\Rightarrow \left(\frac{dt}{ds}\right)^2\Big[\xdd_t + \h_2(\x_t\xd_t)\Big] \\ \nonumber
    &\ \ \ \ \ \ \ \ \ \ \ \ \ 
    + \left(\frac{d^2t}{ds^2} + \alpha(s)\frac{dt}{ds}\right)\xd_t = \zero.
\end{align}
Since $\frac{d^2t}{ds^2} + \alpha(s)\frac{dt}{ds} = 0$ is an ordinary second-order differential equation it has a unique solution for every initial condition $\big(t(0), \frac{dt}{ds}(0)\big)$. Under any of those solutions, the second term vanishes and we have $\xdd_t + \h_2(\x_t, \xd_t) = \zero$ (since $\frac{dt}{ds}\neq 0$). Therefore, under such a time reparameterization, $\x_t(t)$ is a generating solution.

Moreover, since $\xd_s = \frac{dt}{ds}\xd_t$ the initial condition $\frac{dt}{ds}(0)$ defines the initial velocity $\xd_t(0) = \left(\frac{dt}{ds}(0)\right)^{-1}\xd_s(0)$ which uniquely defines the generating solution moving in direction $\widehat{\vv} = \widehat{\xd}_s(0)$. Since any initial speed can be generated this way, every time reparameterization of this sort maps to a corresponding generating solution by initial  conditions and every generating solution can be created by such a time reparameterization. Therefore, there is a bijection between time reparameterizations solving $\frac{d^2t}{ds^2} + \alpha(s)\frac{dt}{ds} = 0$ and generating solutions with initial conditions $(\x, \gamma\widehat{\vv})$.
\end{proof}
\fi

The proof of the above theorem 
\compressedversiononly
(see the expanded version \cite{finslerGeometryForRoboticsArXiv2020})
\fi
shows that any time reparameterization $t(s)$ solving
\begin{align} \label{eqn:ReparameterizationEquation}
    \frac{d^2t}{ds^2} + \alpha(s)\frac{dt}{ds} = 0
\end{align}
induces a generating solution. Since solutions to Equation~\ref{eqn:ReparameterizationEquation} are defined by their initial conditions, given a geometric solution $\x_s$, we can choose a time reparameterization $s(t)$ (the inverse of $t(s)$) so that $\x_t = \x_s\big(s(t)\big)$ is a generating solution whose velocity matches the geometric solution's velocity at a given time $\bar{s}$ (choose $t(s)$ solving Equation~\ref{eqn:ReparameterizationEquation} using initial conditions $t(\bar{s}) = \bar{s}$ and $\frac{dt}{ds}(\bar{s}) = 1$). 
That mapping from $s$ to instantaneous velocity-matching generating solution is smooth, so we can view the geometric solution as {\em smoothly moving between generating solutions}, using the redundant accelerations $\alpha(s)\xd_s$ to do so by speeding up and slowing down along the direction of motion.  

To illustrate path consistency, we designed a geometry, $\h_2(\q, \qd)$, that naturally produces particle paths that avoid a circular object in coordinates, $\q \in \mathbb{R}^{2}$, as 
\begin{align} \label{eqn: phi2_case1}
    \h_2(\q,\qd) = \lambda \| \qd \| ^2 \: \partial_\q \psi\big(\phi(\q)\big)
\end{align}
where $\phi(\q)$ is a differentiable map that captures the distance to a circular object and $\lambda \in \mathbb{R}^+$ is a scaling gain. More specifically, $\phi(\q) = \frac{\|\q - \q_o\| - r}{r}$, where $\q_o$ and $r$ are the circle's center and radius, respectively. Furthermore, $\psi\big(\phi(\q)\big) \in \mathbb{R}^+$ is a barrier potential function, $\psi\big(\phi(\q)\big) = \frac{k}{\phi(\q)^2}$, where $k \in \mathbb{R}^+$ is a scaling gain. Altogether, $\partial_\q \psi\big(\phi(\q)\big)$ produces an increasing repulsive force as distance to the object decreases, and $\| \qd \| ^2$ makes $\h_2(\q, \qd)$ homogeneous of degree 2 in $\qd$. For this experiment, $\lambda = 0.7$, $k = 0.5$ for two scenarios: 1) $\alpha = 1.5$, and 2) $\alpha = 0.75$ for the initial conditions, $(\q_0, \alpha \hat{\qd}_0)$. Eleven vertically spaced particles that follow the above geometry are initialized with the two different initial speeds. Traced paths at the two different speeds are overlaid as shown in Fig. \ref{fig:geometric_consistency}. Noticeably, the paths generated are completely overlapping confirming path consistency.
\section{Finsler geometry} \label{sec:FinslerGeometry}

Here, we derive a broad class of nonlinear geometries, known as Finsler geometries \cite{baoRiemanFinslerGeometry2000}. These geometries arise from Calculus of Variations problems \cite{CalculusOfVariationsGelfandFomin63} that generalize the notion of {\em arc length} to cases where length elements can vary by direction (Minkowski norms \cite{something}). This section builds to a fundamental result showing that applying the Euler-Lagrange equation to these generalized length elements produces a Geometric equation (a Finsler geometry of paths) whose corresponding generator can be derived by applying the Euler-Lagrange equation to the length element's {\em energy form}. Solutions to the latter conserve the derived energy and can thus be viewed as energy levels. And from the above analysis, we can characterize geometric solutions to Finsler geometries as smoothly transitioning between these concretely defined energy levels (generating solutions) by speeding up and slowing down along the direction of motion while remaining along the same common path.


\subsection{The Euler-Lagrange equation: a review}

For completeness we review some background on the Calculus of Variations \cite{CalculusOfVariationsGelfandFomin63}. The Calculus of Variations studies extremal trajectory problems. Given a function $\Lag(\x, \xd)$ of position and velocity, known as a Lagrangian, and the class of smooth trajectories $\Xi$ ranging between two end points $\x_0$ and $\x_1$, we can ask which of those trajectories minimizes the ``total Lagrangian'' across the trajectory:
\begin{align}
    \min_{\x(t)\in\Xi}\int_0^T \Lag(\x, \xd)dt,
\end{align}
where $T$ is understood to vary per trajectory based on its natural time interval. 
The integral $A[\x] = \int \Lag(\x, \xd)dt$ is known as the Lagrangian's {\em action} functional.

\fullversiononly
\begin{figure}[!t]
  \centering
  \includegraphics[width=.95\linewidth]{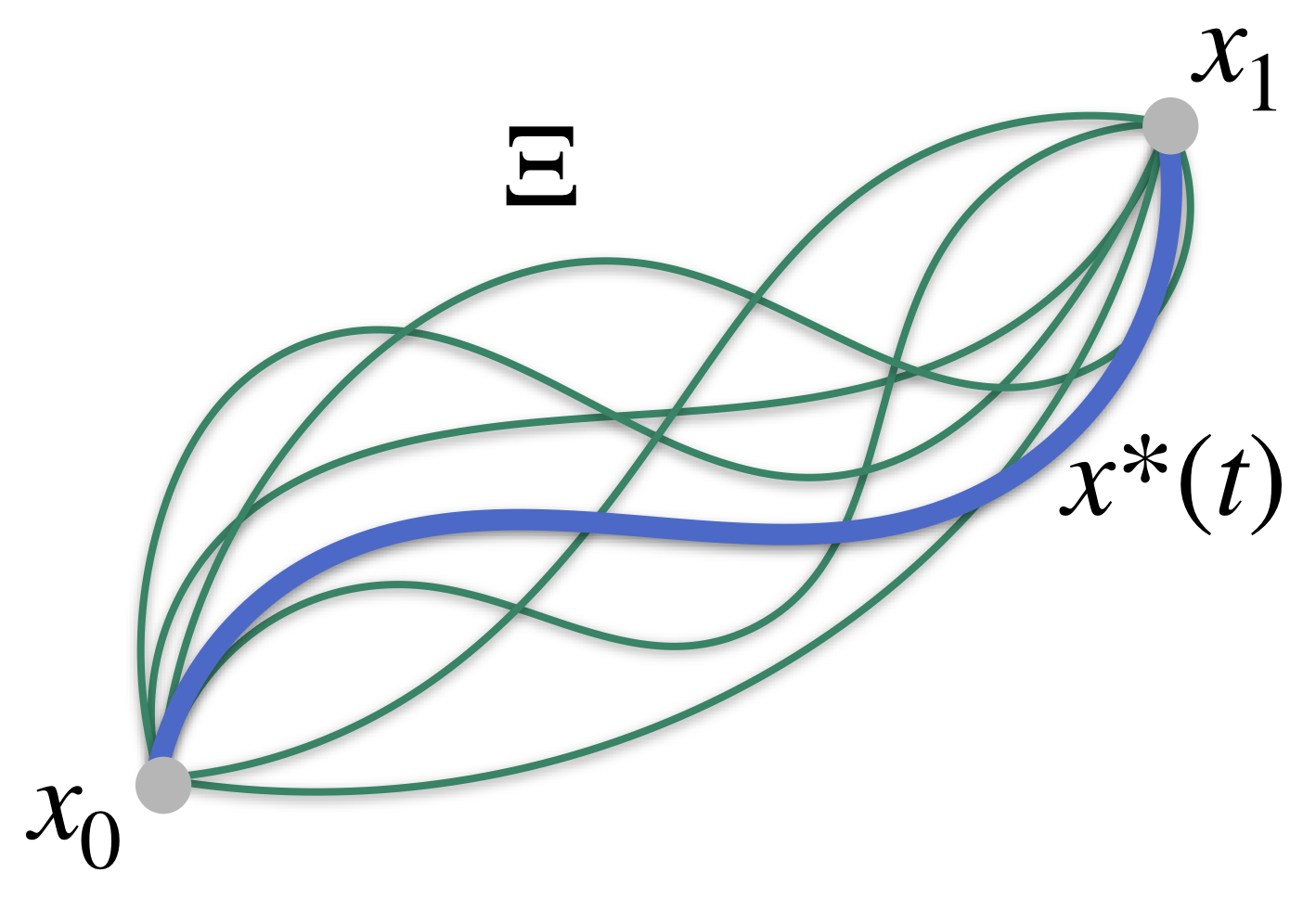}
  \vspace{-4mm}
  \caption{A depiction of a Calculus of Variations problem. Green curves are select trajectories $\x(t)\in\Xi$ between $\x_0$ and $\x_1$ from a smooth family of trajectories $\Xi$; the blue curve $\x^*(t)$ depicts an extremal trajectory that optimizes an objective functional.}
  \label{fig:CalculusOfVariations}
\end{figure}

Figure~\ref{fig:CalculusOfVariations} depicts this problem pictorally. Here the dark trajectories are possible candidates in $\Xi$ and the dotted trajectory is the extremum. 
\fi 

We won't derive it here, but extremal solutions are characterized by solutions to the following boundary-valued second-order differential equation:
\begin{align}
    \frac{d}{dt}\partial_\xd\Lag - \partial_\x\Lag = \zero
    \ \ \mathrm{s.t.}\ \ 
    \left\{
    \begin{array}{l}
        \x(0) = \x_0 \\
        \x(T) = \x_1
    \end{array}
    \right.
\end{align}
This equation $\frac{d}{dt}\partial_\xd\Lag - \partial_\x\Lag = \zero$ is important in its own right and is known as the Euler-Lagrange equation.\footnote{We write it in negated form here relative to the common expression from the Calculus of Variations \cite{CalculusOfVariationsGelfandFomin63} to match better with the equations of motion below, as we'll see.} By the theory of ordinary differential equations \cite{LeeSmoothManifolds2012}, this equation has unique initial-value solutions when it is point-wise well-formed for $\xd\neq\zero$. (We'll see that this means its velocity Hessian $\partial^2_{\xd\xd}\Lag$ is invertible for $\xd\neq\zero$. This won't be the case for Finsler structures (and we'll get the solution redundancy characteristic of geometric equations from that), but it will be the case for the corresponding energy-form generating equation.) This uniqueness of solution means that for any $(\x_0, \xd_0)$, we can play forward the Euler-Lagrange equation uniquely to generate a solution trajectory starting from that position and velocity. We can connect this solution back to the extremal problem by noting that every point $\x_1$ encountered along this initial-value solution is a possible end point, and the solution to the initial-valued problem solves the boundary value problem with boundary constraints $\x_0$ and $\x_1$. In other words, every subtrajectory of the initial-value solution is an extremal solution between its end points. Importantly, this enables us to consider the Euler-Lagrange equation in isolation and understand all of its solutions as extremal solutions of the action.

Expanding the time derivative brings the Euler-Lagrange equation to a more concrete form and clarifies its role as a second-order dynamical system:
\begin{align}
    &\ \ \frac{d}{dt}\partial_\xd\Lag - \partial_\x\Lag = \zero\\
    &\Rightarrow \partial^2_{\xd\xd}\Lag\xdd + \partial_{\xd\x}\Lag\xd - \partial_\x\Lag = \zero \\
    &\Rightarrow \M_\Lag \xdd + \f_\Lag = \zero,
\end{align}
where $\M_\Lag = \partial^2_{\xd\xd}\Lag$ plays a role analogous to a mass matrix and $\f_\Lag = \partial_{\xd\x}\Lag\xd - \partial_\x\Lag$ is a force-like object. Solutions to the Euler-Lagrange equation can be easily integrated forward from an initial position $\x$ and velocity $\xd$ using the solved acceleration form $\xdd = -\M_\Lag^{-1}\f_\Lag$. Note that this solved form is only well-defined when $\M_\Lag$ is invertible as noted earlier. While here we consider $\M_\Lag$ and $\f_\Lag$ as merely {\em analogous} to mass and force, we will see that this analogy takes on a deeper, more concrete, meaning under Finsler geometry, where we require the Lagrangian to take on a special form so that $\Lag(\x, \xd)$, for fixed $\x$, intuitively becomes a squared norm like measure on velocities, giving it an interpretation of ``length squared''. We will see below that, under these particular Lagrangians, $\M_\Lag$ plays a role of mass, $\f_\Lag$ plays a role of force, and the equation $\xdd + \M_\Lag^{-1}\f_\Lag = \zero$, or $\xdd + \h_2(\x, \xd) = \zero$ with $\h_2 = \M_\Lag^{-1}\f_\Lag$, is a geometry generator.

\subsection{Finsler structures}

We call a Lagrangian whose Euler-Lagrange equation can be expressed in the standard geometric form 
\begin{align}
    \mP^\perp_\xd\Big[\xdd + \h_2(\x, \xd)\Big] = \zero,
\end{align}
where $\h_2$ is homogeneous of degree 2 in velocity (HD2), a {\em geometric Lagrangian}. Recall that such geometries described by these geometric equations exhibit an invariance to time-reparameterization (see Section~\ref{sec:GeneralNonlinearGeometry}) with solutions characterizing a {\em geometry of paths}. A {\em Finsler structure} is a particular geometric Lagrangian $\Lag_g$ with the following nice properties:
\begin{enumerate}
    \item $\Lag_g(\x, \xd)\geq 0$ with equality if and only if $\xd = \zero$.
    \item $\Lag_g$ is positively homogeneous (HD1) in $\xd$ so that $\Lag_g(\x, \lambda\xd) = \lambda \Lag_g(\x, \xd)$ for $\lambda\geq 0$.
    \item $\partial^2_{\xd\xd}\Lag_e$ is invertible when $\xd \neq \zero$, where $\Lag_e = \frac{1}{2}\Lag_g^2$. \label{def:FinslerEnergy}
\end{enumerate}
We call $\Lag_e = \frac{1}{2}\Lag_g^2$ defined in Property~\ref{def:FinslerEnergy}, the corresponding Finsler {\em energy}. 

The positive homogeneity requirement enforces that the action functional is natively independent of time reparameterization. A time reparameterization defines $s = s(t)$ with $\xd_s = \frac{dt}{ds}\xd_t$, so homogeneity implies the action functional has the property
\begin{align*}
    A[\x_s]
    &= \int \Lag_g(\x_s, \xd_s)ds = \Lag_g\left(\x_t, \frac{dt}{ds}\xd_t\right)ds\\
    &= \int\Lag_g(\x_t, \xd_t)\frac{dt}{ds} ds \\
    &= \int\Lag_g(\x_t, \xd_t)dt = A[\x_t].
\end{align*}
This property suggests that {\em solutions} to the Euler-Lagrange equation of $\Lag_g$ must also be independent of time reparameterization, i.e. they form paths. Theorem~\ref{thm:FundamentalTheoremOfFinslerGeometry} proves that conjecture and concretely connects Finsler geometries to the generalized notion of nonlinear geometries of paths outlined above.
The speed independence of this action functional, in conjunction with the conditions of a Finsler structure listed above, also suggest we can view the Finsler structure $\Lag_g$ as a generalized {\em length element}. The action functional is, therefore, a generalized arc-length integral.

Since the Finsler structure $\Lag_g$ is HD1, it's associated Finsler energy has the property $\Lag_e(\x, \lambda\xd) = \frac{1}{2}\big(\Lag_g(\x, \lambda\xd)\big)^2 = \lambda^2\frac{1}{2}\big(\Lag_g(\x, \xd)\big)^2 = \lambda^2\Lag_e(\x, \xd)$. That means the Finsler energy $\Lag_e$ is positively homogeneous of degree 2 (HD2). A useful property homogeneity is given by Euler's theorem on homogeneous functions \cite{lewisHomogeneousFunctions1969}, which states that if $\Lag(\x, \xd)$ is homogeneous of degree $k$ in $\xd$, then the Hamiltonian, a conserved quantity under the Euler-Lagrange equation \cite{LeonardSusskindClassicalMechanics}, is
\begin{align}
    \Ham_\Lag = \partial_\xd\Lag^\tr\xd - \Lag = (k-1)\Lag.
\end{align}
In the case of the Finsler energy $\Lag_e = \frac{1}{2}\Lag_g^2$, we have $k = 2$, so
\begin{align}
    \Ham_{\Lag_e} = (k-1)\Lag_e  = \Lag_e.
\end{align}
The equations of motion under the Finsler energy $\Lag_e$, which we call the energy equations, are 
\begin{align}
    &\partial^2_\xd\Lag_e\xdd + \partial_{\xd\x}\Lag_e\xd - \partial_x\Lag_e = \zero \\
    &\ \ \ \Leftrightarrow \M_e\xdd + \f_e = \zero,
\end{align}
where $\M_e = \partial^2_\xd\Lag_e$ and $\f_e = \partial_{\xd\x}\Lag_e\xd - \partial_x\Lag_e$. Since these equations are known to conserve $\Ham_e$, and in this case $\Ham_e = \Lag_e$, we see that this Finsler energy is conserved. $\Lag_e$ is often referred to as the {\em energy} of the system, and $\M_e$ is its {\em energy tensor}.

Note that the third requirement on Finsler structures given above is actually a requirement on the Finsler {\em energy} $\Lag_e$. It ensures that $\M_e$ is invertable by definition when $\xd \neq \zero$. The equations of motion $\M_e\xdd + \f_e = \zero$, therefore, can always be solved to give a unique acceleration form
\begin{align}
    \xdd + \M_e^{-1}\f_e = \zero.
\end{align}
This equation, derived from the Finsler {\em energy} $\Lag_e$, is known as the {\em geodesic equation}, and we will see that it acts as a generator for the geometry expressed by the Finsler {\em structure} $\Lag_g$'s equations of motion.

Without proving it here, we note that derivatives reduce a function's homogeneity by 1 (a general property of homogeneous functions). Therefore, examining $\f_e$, we see
\begin{align}
    \f_e = \partial_{\xd\x}\Lag_e\xd - \partial_\x\Lag_e.
\end{align}
Here $\partial_\x\Lag_e$ is already HD2, and $\xd$ multiplies the HD1 $\partial_{\xd\x}\Lag_e$ making $\partial_{\xd\x}\Lag_e\xd$ HD2 as well. So $\f_e$ is HD2 in its entirety. Moreover, $\M_e = \partial^2_{\xd\xd}\Lag_e$ has two derivatives so it is HD0 (i.e. $\M_e(\x, \lambda\xd) = \M_e(\x, \xd)$, meaning the energy tensor is independent of the {\em scale} of the velocity, depending only on $\xd$'s {\em directionality}). That means $\h_2(\x, \xd) = \M_e^{-1}\f_e$ is HD2, making the geodesic equation $\xdd + \M_e^{-1}\f_e = \zero$ a generating equation with associated geometric equation
\begin{align}
    P^{\perp}_\xd\Big[\xdd + \M_e^{-1}\f_e\Big] = \zero.
\end{align}
The following theorem shows that this geometric equation is precisely that characterized by $\Lag_g$'s equations of motion. We denote the {\em geometric} equations of motion by $\M_g \xdd + \f_g = \zero$ where $\M_g = \partial^2_{\xd\xd}\Lag_g$ and $\f_g = \partial_{\xd\x}\Lag_g - \partial_\x\Lag_g$ for Finsler {\em structure} $\Lag_g$. In this case, $\M_g$ is reduced rank, so we cannot solve for a unique acceleration. Instead, the redundancy is precisely that expressed by the geometric equation.

\begin{theorem} \label{thm:FundamentalTheoremOfFinslerGeometry}
Let $\Lag_g$ be a Finsler structure with energy form $\Lag_e = \frac{1}{2}\Lag_g^2$. Then the energy equation $\M_e\xdd + \f_e = \zero$ is a generating equation $\xdd + \h_2(\x,\xd) = \zero$ with $\h_2 = \M_e^{-1}\f_e$. The associated geometric equation $\mP_\xd\big[\xdd + \M_e^{-1}\f_e\big] = \zero$ is given by the geometric equations of motion $\M_g\xdd + \f_g = \zero$.
\end{theorem}

\begin{proof}
We already observed that $\h_2(\x, \xd) = \M_e^{-1}\f_e$ is homogeneous of degree 2, so $\xdd + \h_2(\x, \xd) = \zero$ is a generating equation where it is (uniquely) defined. Since $\M_e$ is invertible by definition when $\xd\neq\zero$, it is only undefined for $\xd = \zero$. But for $\xd = \zero$, solutions to $\M_e\xdd + \f_e = \zero$ are stationary point trajectories ($\xdd = \zero$) since $\f_e$ is homogeneous. Therefore, defining $\h_2(\x, \zero) = \zero$ creates matching limiting behavior (independent of the characteristic properties of generating equations, which characterize geometrically consistent generating trajectories for $\xd\neq\zero$), so $\xdd + \h_2(\x, \xd) = \zero$ is a generating equation with solutions consistent with $\M_e\xdd + \f_e = \zero$. The energy equation, therefore, generates a geometry. We will see below that this geometry is given by the geometric equations of motion.

To show $\M_g\xdd + \f_g = \zero$ is a geometric equation, we calculate explicit expressions for $\M_g$ and $\f_g$. Since $\Lag_e = \frac{1}{2}\Lag_g^2$ and $\Lag_g = (2\Lag_e)^{\frac{1}{2}}$, we have 
\begin{align} \label{eqn:GeometricMassExpansion}
    \nonumber
    \M_g
    &= \partial^2_{\xd\xd}\Lag_g 
    = \sqrt{2}\partial_\xd\left[\partial_\xd\Lag_e^{\frac{1}{2}}\right]
    = \sqrt{2}\partial_\xd\left(\frac{1}{2}\Lag_e^{-\frac{1}{2}}\partial_\xd\Lag_e\right)\\\nonumber
    &= \frac{\sqrt{2}}{2}\left[
        \left(-\frac{1}{2}\right)\Lag_e^{-\frac{1}{2} - 1} \partial_\xd\Lag_e\partial_\xd\Lag_e^\tr
        + \Lag_e^{-\frac{1}{2}}\partial^2_{\xd\xd}\Lag_e
    \right] \\
    &= \frac{1}{(2\Lag_e)^{\frac{1}{2}}}\left[
        \partial^2_{\xd\xd}\Lag_e - \frac{1}{2\Lag_e}\partial_\xd\Lag_e\partial_\xd\Lag_e^\tr
    \right] \\\nonumber
    &= \frac{1}{\Lag_g}\left[\M_e - \frac{\p_e\p_e^\tr}{\p_e^\tr\M_e^{-1}\p_e}\right],
\end{align}
where we use $\M_e = \partial^2_{\xd\xd}\Lag_e$, denote $\p_e = \partial_\xd\Lag_e$ (this quantity is called the {\em generalized momentum} and has a recurring role in Finsler theory), and use the following identity, $\Lag_e = \frac{1}{2}\p_e^\tr\M_e^{-1}\p_e$ (see Lemma~\ref{lma:MomentumIdentities}).
Denoting $\mR_\xd = \M_e - \frac{\p_e\p_e^\tr}{\p_e^\tr\M_e^{-1}\p_e}$ and noting $\p_e = \M_e\xd$ (see again Lemma~\ref{lma:MomentumIdentities}), we see that $\M_g = \frac{1}{\Lag_g}\mR_\xd$ is a reduced rank matrix with null space spanned by $\xd$ since
\begin{align}
    \mR_\xd\xd 
    &= \left(\M_e - \frac{\p_e\p_e^\tr}{\p_e^\tr\M_e^{-1}\p_e}\right) \M_e^{-1}\p_e \\
    &= \M_e\M_e^{-1}\p_e - \frac{\p_e\big(\p_e^\tr\M_e^{-1}\p_e\big)}{\p_e^\tr\M_e^{-1}\p_e} \\
    &= \p_e - \p_e = \zero.
\end{align}
Using a calculation analogous to that which we used for Equation~\ref{eqn:GeometricMassExpansion} we get
\begin{align}
    \partial_{\xd\x}\Lag_g = \frac{1}{\Lag_g}\left[
        \partial_{\xd\x}\Lag_e - \frac{1}{2\Lag_e}\partial_\xd\Lag_e\partial_\x\Lag_e^\tr
    \right],
\end{align}
and separately,
\begin{align}
    \partial_{\x}\Lag_g = \partial_\x(2\Lag_e)^\frac{1}{2}
    = \frac{\sqrt{2}}{2}\Lag_e^{-\frac{1}{2}}\partial_\x\Lag_e
    = \frac{1}{\Lag_g}\partial_\x\Lag_e,
\end{align}
so combining we get
\begin{align}
    \f_g 
    &= \partial_{\xd\x}\Lag_g\xd - \partial_\x\Lag_g \\
    &= \frac{1}{\Lag_g}\left[
        \big(\partial_{\xd\x}\Lag_e\xd - \partial_\x\Lag_e\big)
        - \frac{1}{2\Lag_e}\partial_\xd\Lag_e\partial_\x\Lag_e^\tr\xd
    \right]\\\label{eqn:fgExpression}
    &= \frac{1}{\Lag_g}\left[\f_e - \frac{1}{2\Lag_e}\partial_\xd\Lag_e\partial_\x\Lag_e^\tr\xd\right].
\end{align}
We first show that $\xd^\tr\f_g = \zero$ (i.e. $\f_e$ is orthogonal to $\xd$) and then use that insight to derive an explicit projected expression for $\f_g$:
\begin{align}
    \xd^\tr\f_g 
    &= \frac{1}{\Lag_g}\left[
        \xd^\tr\f_e - \frac{1}{2\Lag_e}\xd^\tr\partial_\xd\Lag_e\partial_\x\Lag_e^\tr\xd
    \right] \\
    &= \frac{1}{\Lag_g}\bigg[
        \xd^\tr\big(\partial_{\xd\x}\Lag_e\xd - \partial_\x\Lag_e\big)\\\nonumber
    &\ \ \ \ \ \ \ \ \ \ \ \ \ \ \ \ \ \ \ 
    - \frac{\frac{1}{2}\xd^\tr\partial_\xd\Lag_e}{\Lag_e}\partial_x\Lag_e^\tr\xd
    \bigg].
\end{align}
From Lemma~\ref{lma:MomentumIdentities} we also get $\Lag_e = \frac{1}{2}\p_e^\tr\xd = \frac{1}{2}\partial_\xd\Lag_e^\tr\xd$, so the expression reduces to
\begin{align}
    \xd^\tr\f_g = \frac{1}{\Lag_g}\left[
        \xd^\tr\partial_{\xd\x}\Lag_e\xd - 2\partial_\x\Lag_e^\tr\xd
    \right].
\end{align}
Finally, 
\begin{align}
    &\xd^\tr\partial_{\xd\x}\Lag_e\xd - 2\partial_\x\Lag_e^\tr\xd \\
    &\ \ = \left(\xd^\tr\partial_\x\big(\partial_\xd\Lag_e\big)
        - 2\partial_\x\Lag_e
    \right)\xd \\
    &\ \ = \partial_\x\left(\xd^\tr\partial_\xd\Lag_e - 2\Lag_e\right)\xd \\
    &\ \ = \partial_\x\left(2\Lag_e - 2\Lag_e\right)\xd = \zero,
\end{align}
again since $\Lag_e = \frac{1}{2}\partial_\xd\Lag_e^\tr\xd$ by Lemma~\ref{lma:MomentumIdentities}. Therefore, $\xd^\tr\f_g = \zero$.


By Equation~\ref{eqn:fgExpression} $\f_g$ takes the form $\f_g = \frac{1}{\Lag_g}\left[\f_e - \alpha\p_e\right]$ (with $\p_e = \partial_\xd\Lag_e$ and $\alpha = \frac{1}{2\Lag_e}\partial_\x\Lag_e^\tr\xd$). 
Since we just saw that $\f_g$ is orthogonal to $\xd$, 
that $\alpha$ coefficient must be precisely the coefficient on $\p_e$ needed to remove the component of $\f_e$ along $\xd$. Explicitly, $\alpha$ must satisfy $\xd^\tr\big[\f_e - \alpha\p_e\big] = \zero$, which means
\begin{align}
    \alpha = \frac{\xd^\tr\f_e}{\xd^\tr\p_e}
\end{align}
is another expression for $\alpha$. That means
\begin{align}
    \f_g 
    &= \frac{1}{\Lag_g}\big[\f_e - \alpha\p_e\big]
    = \frac{1}{\Lag_g}\left[\f_e - \left(\frac{\xd^\tr\f_e}{\xd^\tr\p_e}\right)\p_e\right]\\
    &= \frac{1}{\Lag_g}\left[\I - \frac{\p_e\xd^\tr}{\xd^\tr\p_e}\right]\f_e.
\end{align}
Noting again that $\p_e = \M_e\xd$ (see Lemma~\ref{lma:MomentumIdentities}), we have 
\begin{align}
    \f_g 
    &= \frac{1}{\Lag_g}\left[
        \I - \frac{\M_e\xd\xd^\tr}{\xd^\tr\M_e\xd}
    \right] \f_e \\
    &= \frac{1}{\Lag_g}\M_e\left[
        \M_e^{-1} - \frac{\xd\xd^\tr}{\xd^\tr\M_e\xd}
    \right]\f_e \\
    &= \frac{1}{\Lag_g}\M_e\mR_{\p_e}\f_e,
\end{align}
where
\begin{align}
    \mR_{\p_e} = \M_e^{-1} - \frac{\xd\xd^\tr}{\xd^\tr\M_e\xd}
\end{align}
is a reduced rank matrix analogous to $\mR_\xd$, with null space spanned by $\p_e = \M_e\xd$.

Combining all of these expressions so far, we get geometric equations of motion
\begin{align}
    \M_g\xdd + \f_g 
    = \frac{1}{\Lag_g}\mR_\xd\xdd + \frac{1}{\Lag_g}\M_e\mR_{\p_e}\f_e = \zero.
\end{align}
However, $\mR_\xd$ and $\mR_{\p_e}$ are related by the identity $\M_e\mR_{\p_e}\M_e = \mR_\xd$ since
\begin{align}
    \mR_\xd 
    &= \M_e - \frac{\p_e\p_e^\tr}{\p_e^\tr\M_e^{-1}\p_e} \\
    &= \M_e - \frac{\M_e\xd\xd^\tr\M_e}{\big(\M_e\xd\big)^\tr \M_e^{-1} \big(\M_e\xd\big)}\\
    &= \M_e\left[
        \M_e^{-1} - \frac{\xd\xd^\tr}{\xd^\tr\M_e\xd}
    \right] \M_e \\
    &= \M_e\mR_{\p_e}\M_e.
\end{align}
Therefore, the geometric equations of motion can be expressed
\begin{align}
    &\frac{1}{\Lag_g}\M_e\mR_{\p_e}\M_e\xdd + \frac{1}{\Lag_g}\M_e\mR_{\p_e}\f_e = \zero \\
    &\Rightarrow \frac{1}{\Lag_g}\M_e\mR_{\p_e}\big[\M_e\xdd + \f_e\big] = \zero \\
    &\mathrm{or}\ \mP_{\p_e}^\perp\big[\M_e\xdd + \f_e\big] = \zero,
\end{align}
where $\mP_{\p_e}^\perp$ is an orthogonal projector with null space $\p_e$ (any matrix with $\p_e$ spanning its null space would express the same equation). Solutions to this equation are exactly those for which $\M_e\xdd + \f_e$ lies along the null space $\p_e$, so they can be expressed as solutions to the unprojected equation
\begin{align}
    \M_e\xdd + \f_e + \alpha\p_e = \zero
\end{align}
for any $\alpha\in\R$. Since $\p_e = \M_e\xd$ by Lemma~\ref{lma:MomentumIdentities}, those solutions also solve
\begin{align}
    &\ \ \M_e\xdd + \f_e + \alpha\M_e\xd = \zero \\
    &\Rightarrow\ \ \xdd + \M_e^{-1}\f_e + \alpha\xd = \zero
\end{align}
for any $\alpha\in\R$, which are exactly the solutions to 
\begin{align}
    \mP_\xd^\perp\big[\xdd + \M_e^{-1}\f_e\big] = \zero,
\end{align}
where $\mP_\xd^\perp$ is the orthogonal projector with null space spanned by $\xd$. This equation is the geometric equation induced by generator $\xdd + \M_e^{-1}\f_e = \zero$. Therefore, the energy equations of motion $\M_e\xdd + \f_e = \zero$ form a geometric generator whose geometric equation is given by the geometric equations of motion $\M_g\xdd + \f_g = \zero$.
\end{proof}

\fullversiononly
The above proof alludes to a number of properties of the energy $\Lag_e$ and its generalized momentum $\p_e = \partial_\xd\Lag_e$. These results are collected and proved in the following Lemma. Note that the following identities match those seen in Riemannian geometry simply with the Riemannian metric replaced by the generalized metric tensor. See Section~\ref{sec:RiemannianGeometry} for details.
\else 
The required properties of Finsler structures enable us to prove (see \cite{finslerGeometryForRoboticsArXiv2020}) some useful identities relating generalized momentum and energies. These identities match the form seen in Riemannian geometry with the Riemannian metric replaced by the generalized metric tensor $\M_e = \partial^2_{\xd\xd}\Lag_e$.
\fi 

\begin{lemma}[Energy and momentum identities]\label{lma:MomentumIdentities}
Let $\Lag_g$ be a Finsler structure with energy form $\Lag_e = \frac{1}{2}\Lag_g^2$, and denote the generalized momentum $\p_e = \partial_\xd\Lag_e$. Then 
\begin{enumerate}
    \item $\p_e = \M_e\xd$
    \item $\Lag_e = \frac{1}{2}\p_e^\tr\xd = \frac{1}{2}\xd^\tr\M_e\xd = \frac{1}{2}\p_e^\tr\M_e^{-1}\p_e$.
\end{enumerate}
\end{lemma}
\fullversiononly
\begin{proof}
Since $\Lag_e$ is homogeneous of degree 2, as we've seen $\Ham_{\Lag_e} = \p_e^\tr\xd - \Lag_e = \Lag_e$. This directly implies
\begin{align}
    \Lag_e = \frac{1}{2}\p_e^\tr\xd
\end{align}
giving the first form of identity (2). The rest of those identities derive from identity (1); to prove that identity, we take the gradient of this expression for $\Lag_e$
\begin{align}
    \partial_\xd\Lag_e
    &= \partial_\xd\big[\frac{1}{2}\partial_{\xd}\Lag_e^\tr\xd\big] \\
    &= \frac{1}{2}\partial^2_{\xd\xd}\Lag_e\xd + \frac{1}{2}\partial_\xd\Lag_e
\end{align}
which implies
\begin{align}
    \partial^2_{\xd\xd}\Lag_e\xd = 2\partial_\xd\Lag_e - \partial_\xd\Lag_e
    = \partial_\xd\Lag_e
\end{align}
or $\M_e\xd = \p_e$.
\end{proof}
\fi 

\fullversiononly
\section{Riemannian geometry is a Finsler geometry} \label{sec:RiemannianGeometry}

Riemannian geometry is a special case of Finsler geometry. This section shows how many of the most important properties of Riemannian geometry arise from the more general properties of Finsler geometry.
The Riemannian Finsler structure is $\Lag_g = \big(\xd^\tr\G(\x)\xd\big)^{\frac{1}{2}}$, where $\G(\x)$ is a smoothly changing symmetric positive definite matrix. Since $\G$ is independent of $\xd$, it's easy to see that the Finsler structure is HD1
\begin{align*}
    \Lag_g(\x, \lambda\xd)
    &= \big((\lambda\xd)^\tr\G(\x)(\lambda\xd)\big)^{\frac{1}{2}} \\
    &= \big(\lambda^2\xd^\tr\G(\x)\xd\big)^{\frac{1}{2}}\\
    &= \lambda \Lag_g(\x, \xd).
\end{align*}
Likewise, since $\G$ is positive definite $\Lag_g\geq\zero$ with equality only when $\xd = \zero$. And since $\Lag_e = \frac{1}{2}\xd^\tr\G\xd$, we have $\M_e = \partial^2_\xd\Lag_e = \G$, which is everywhere invertible.
The matrix $\G$ is called a {\em Riemannian metric}, and plays the role of the Finsler {\em metric tensor}.
The Finsler structure defines a norm on the tangent space (the space of velocities at a point $\x$)
\begin{align}
    \Lag_g = 
    (\xd^\tr\G\xd)^\frac{1}{2} = \|\xd\|_\G > 0\ \ \mbox{for $\xd\neq 0$},
\end{align}
showing that the action defining the extremal problem's objective is
\begin{align}
    A[\x] = \int\Lag_g dt = \int\|\xd\|_\G dt.
\end{align}
This action functional can be understood as a generalized arc-length integral across the trajectory.

Since the action is a generalized arc-length integral it seems natural that this measure would be invariant to time-reparameterization of the trajectory (i.e. invariant to speed profile across the trajectory). Using time-reparameterization $t(s)$ with $\xd_s = \frac{dt}{ds}\xd_t$, we can perform the calculation of Section~\ref{sec:RiemannianGeometry} explicitly
\begin{align*}
    A[\x_s] 
    &= \int\|\xd_s\|_\G ds = \int \left\|\frac{dt}{ds}\xd_t\right\|_\G ds \\
    &= \int\|\xd_t\|_\G\frac{dt}{ds}ds = \int\|\xd_t\|_\G dt = A[\x_t].
\end{align*}

Per Theorem~\ref{thm:FundamentalTheoremOfFinslerGeometry} we would expect the Riemannian Finsler structure's Euler-Lagrange equation $\M_g\xdd + \f_g = \zero$ to have a reduced rank $\M_g = \partial^2_\xd\Lag_g$ (otherwise, it would have a unique (non-redundant) solution).
Indeed, by calculation, we have
\begin{align}
    \M_g 
    &= \partial^2_{\xd\xd}\Lag_g = \partial_\xd\left[\partial_\xd(\xd^\tr\G(\x)\xd)^{\frac{1}{2}}\right] \\
    &= \partial_\xd\left[\frac{1}{2}(\xd^\tr\G\xd)^{-\frac{1}{2}} 2\G\xd\right]
    = \partial_\xd\left[\frac{\G\xd}{\|\xd\|_\G}\right] \\
    &= \frac{\|\xd\|_\G\G - \G\xd\left(\frac{\xd^\tr\G}{\|\xd\|_\G}\right)}{\|\xd\|_\G^2} \\
    &= \frac{1}{\|\xd\|_\G}\left[\G - \frac{\G\xd\xd^\tr\G}{\|\xd\|^2_\G}\right] \\
    &= \frac{1}{\|\xd\|_\G} \G^{\frac{1}{2}}\left[\I - \widehat{\vv}\widehat{\vv}^\tr\right] \G^{\frac{1}{2}},
\end{align}
where $\vv = \G^{\frac{1}{2}}\xd$ and $\widehat{\vv} = \frac{\vv}{\|\vv\|}$. This matrix is reduced rank since $\I - \widehat{\vv}\widehat{\vv}^\tr$ is an orthogonal projector (with null space spanned by $\vv$).
$\M_g$'s null space is, therefore, spanned by $\uu$ such that $\G^{\frac{1}{2}}\uu = \vv$. Since $\vv = \G^\frac{1}{2}\xd$, it must be that $\uu=\xd$. Therefore, solutions to $\M_g\xdd + \f_g = \zero$ can be expressed as any nominal solution $\xdd_0$ offset by a null space element $\alpha \xd$, i.e. $\xdd = \xdd_0 + \alpha\xd$. 

Remember that the Hamiltonian (conserved quantity) of the energy system given in this case by $\Lag_e = \frac{1}{2}\Lag_g^2$ is (generically) $\Ham_{\Lag_e} = \partial_\xd\Lag_e^\tr\xd - \Lag_e$ In the case of Riemannian geometry, this Hamiltonian is known to evaluate to
\begin{align}
    \Ham_{\Lag_e} = \big(\G\xd\big)^\tr\xd - \frac{1}{2}\xd^\tr\G\xd
    = \frac{1}{2}\xd^\tr\G\xd = \Lag_e.
\end{align}
From the above discussion around homogeneous functions (see Section~\ref{sec:GeneralNonlinearGeometry}) we  see that this seemingly coincidental results actually derives from the second-degree homogeneity of $\Lag_e$, which in turn derives from the first-degree homogeneity of the Finsler structure $\Lag_g = \big(\xd^\tr\G(\x)\xd\big)^\frac{1}{2}$, i.e. the Riemannian length element. 

The energy equations $\M_e\xdd + \f_e = \zero$ in Riemannian geometry have 
\begin{align}
    \M_e &= \partial^2_\xd \Lag_e = \G\\
    \mbox{and}\ \ \f_e &= \partial_{\xd\x}\Lag_e\xd - \partial_\x\Lag_e \\\nonumber
    &= \partial_\x\big(\G\xd\big)\xd - \partial_\x\left(\frac{1}{2}\xd^\tr\G\xd\right). 
\end{align}
From these expressions we can easily see that $\M_e$ is HD0 and $\f_e$ is HD2, making $\xdd + \M_e^{-1}\f_e = \zero$ an HD2 geometry generator. We do not fully derive the geometric equation here, but this result in conjunction with the above observation that $\M_g$ is reduced rank support the results of Theorem~\ref{thm:FundamentalTheoremOfFinslerGeometry}.


\begin{figure*}[hbt!]
  \vspace{2.5mm}
  \centering
  \includegraphics[width=.99\linewidth]{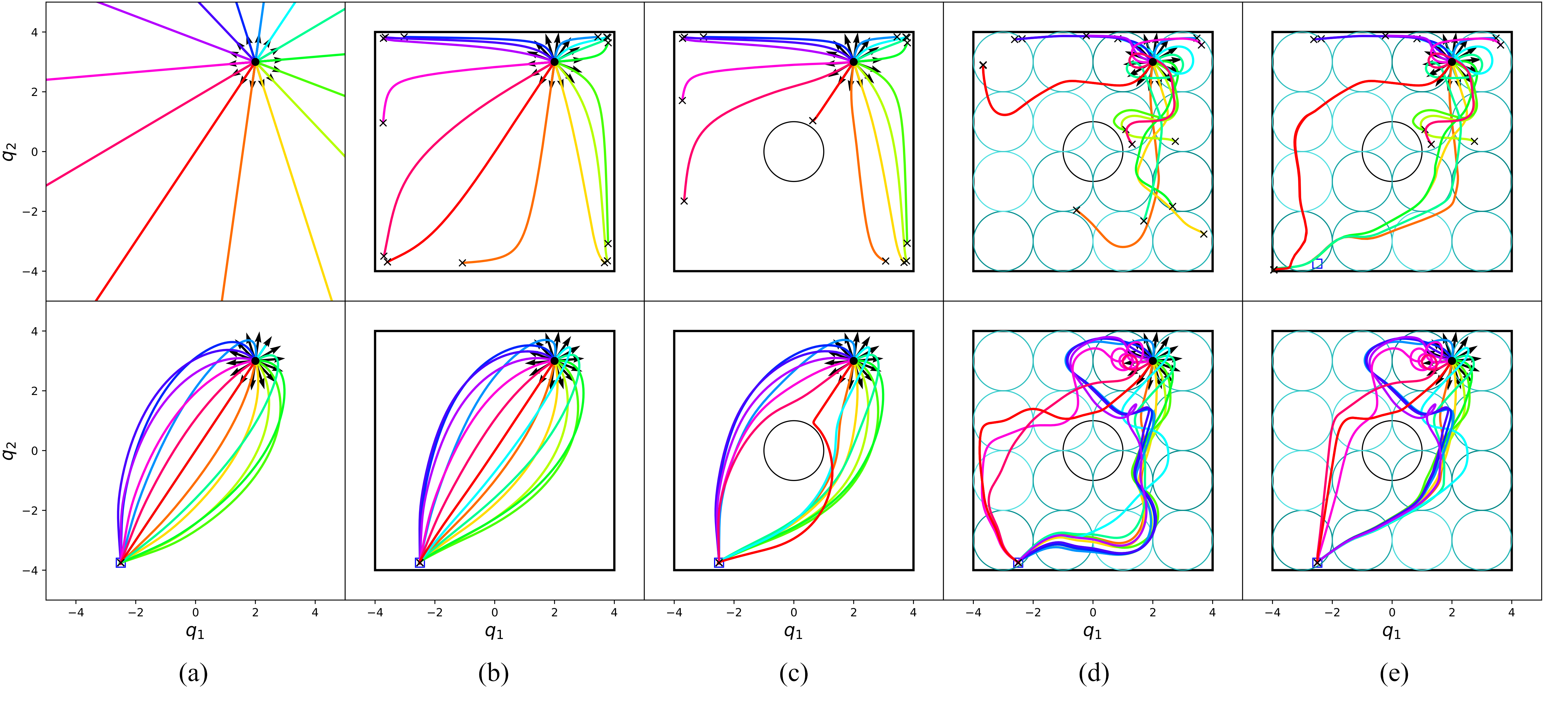}
  \vspace{-5mm}
  \caption{Geometric fabrics, an application of general nonlinear and Finsler geometries for the modeling and design of reactive robotic behavior. Top row, a sophisticated geometry constructed layer-by-layer with increasing complexity from left-to-right. Bottom row, the resulting convergent system when forced by an objective potential. See Section~\ref{sec:GeometricFabrics} for details.}
  \label{fig:geometries_optimization_horizontal}
  \vspace{-4mm}
\end{figure*}

Finally, we note that there is a direct one-to-one correspondence between Riemannian geometric energy systems and classical mechanical systems, a formulation known as geometric mechanics \cite{bullo2004geometric}. Under this correspondence, $\M_e = \G$ is the generalized mass matrix and $\f_e$ captures fictitious forces such as Coriolis and centripetal forces. The energy $\Lag_e = \frac{1}{2}\xd^\tr\G\xd$ is the kinetic energy of the mechanical system and the quantity $\p_e = \partial_\xd\Lag_e = \G\xd$ is the generalized momentum. These quantities match the generalized versions given by Lemma~\ref{lma:MomentumIdentities}.

System solutions of the generator $\M_e\xdd + \f_e = \zero$ conserve the Hamiltonian and are, therefore, constant energy solutions and can be considered to be energy levels. Since $\M_e\xdd + \f_e = \zero$ is a generator, the solutions are geometrically consistent (solutions of all initial value problems with initial velocity pointing in the same direction from the same initial point follow overlapping paths). This geometric property of classical mechanical systems is less often quoted in robotics, but is clear from the generalized development presented here.


\else 

\subsection{Riemannian geometry is a Finsler geometry} \label{sec:RiemannianGeometry}

Riemannian geometry \cite{LeeRiemannianManifolds97} is a type of Finsler geometry (see \cite{finslerGeometryForRoboticsArXiv2020} for more details).
Specifically, the Riemannian Finsler structure is $\Lag_g = \big(\xd^\tr\G(\x)\xd\big)^\frac{1}{2}$ where $\G(\x)$ is the symmetric positive definite Riemannian metric; it can be seen as a generalized length element. The energy form $\Lag_e = \mathcal{K} = \frac{1}{2}\xd^\tr\G(\x)\xd$ matches the  kinetic energy of a classical mechanical system $\G(\x)\xdd + \f_e(\x, \xd) = \zero$, and it is easy to show that $\f_e = \partial_\x\big(\G\xd\big)\xd - \partial_\x\big(\frac{1}{2}\xd^\tr\G\xd\big)$ capture the standard fictitious forces. The Finsler energy tensor is the Riemannian metric $\M_e = \partial^2_\xd\Lag_e = \G$, the generalized momentum is $\p_e = \M_e\xd = \G\xd$, and the energy automatically matches the expression of Lemma~\ref{lma:MomentumIdentities} $\mathcal{K} = \frac{1}{2}\xd^\tr\G(\x)\xd= \frac{1}{2}\xd^\tr\M_e(\x)\xd$. For Riemannian systems, connection between the geometry and this system is known as Geometric Mechanics \cite{bullo2004geometric}.

\fi 

\section{An application: Geometric Fabrics} \label{sec:GeometricFabrics}

One important application of general nonlinear and Finsler geometries is in the design of reactive robotic behavior using {\em geometric fabrics} \cite{geometricFabricsForMotionArXiv2020}, where behavior is modeled as HD2 generalized nonlinear geometries. The second-order homogeneity property models the behavior as a geometry of paths making it independent of speed and giving the system a path consistency that is important for design intuition. These behaviors are designed as a combination of many modular components, each weighed together using the metric tensor of a paired Finsler energy as its weight matrix (a metric-weighted average of component geometries). 
Importantly, each component can be understood as a Finsler geometry, with a well-defined conserved Finsler energy whose metric tensor defines the priority, whose geometry has been {\em bent} to align with the HD2 nonlinear geometry defining the desired behavior. This bending process curves the Finsler geometry in a way that both preserves its geometric HD2 properties and preserves energy conservation (the bending term performs no work on the system and is known as a {\em zero work modification}). What we get in the end is an HD2 geometry that aligns with the desired behavioral geometry but also retains the Finsler geometry's energy metric for prioritization.

Figure~\ref{fig:geometries_optimization_horizontal} depicts a complex geometry in a 2D point space constructed layer-by-layer. From left to right (top row), we see first an underlying Euclidean geometry (first panel) creating straight line behaviors. Adding a barrier geometry curves those geodesic paths to bend away from the walls (second panel), and adding an obstacle barrier (third panel) ensures that the system never hits a circular obstacle as well. Then (fourth panel) a random vortex geometry is added which makes the geodesics bend and twist randomly. Despite the random motions, the resulting geodesics still never penetrate either type of barrier because of the prior layers' contributions. Finally (last panel), an attractor geometry funnels the geodesics heuristically toward a target. In all cases, Finsler energies shape each geometry's priority in the weighted average so it dominates only when most important. See \cite{optimizationFabricsForBehavioralDesignArXiv2020} for an in-depth description the geometries and Finsler energies used in this example.

The theory of geometric fabrics ensures that geometries constructed in this way can be {\em optimized over} to create system {\em goals} represented by local minima of an objective potential simply by forcing the system with the potential's negative gradient. We do just that for each of these geometries (Figure~\ref{fig:geometries_optimization_horizontal}, bottom row) using a simple attractor potential of the type described in \cite{optimizationFabricsForBehavioralDesignArXiv2020}. In each case, the geometry greatly shapes the behavior of the system en route, but all differential equations ultimately converge to the target in the end as guaranteed by the theory of fabrics.


\section{CONCLUSIONS}

Geometric fabrics are powerful tools for designing robot behavior and are an important application of the generalized geometries described here. The derivations here informed the development of Geometric fabrics \cite{geometricFabricsForMotionArXiv2020}, which have produced the most flexible, intuitive, and consistent provably-stable tools to date for designing reactive robot behavior. We hope these derivations and exposition will enable and inspire many more applications of these more general geometries within robotics in the same way Riemannian geometry has already found substantial application.

\compressedversiononly
\clearpage
\fi 
{

\bibliographystyle{IEEEtran}
\compressedversiononly
\bibliography{refs}
\else
\bibliography{refs_aug}
\fi
}

\end{document}